\pdfoutput=1

\documentclass[11pt]{article}

\usepackage{acl}

\usepackage{times}
\usepackage{latexsym}
\usepackage{graphicx}
\usepackage{booktabs}

\usepackage[T1]{fontenc}

\usepackage[utf8]{inputenc}

\usepackage{microtype}

\title{Salient Span Masking for Temporal Understanding}

\author{Jeremy R. Cole \qquad Aditi Chaudhary \qquad Bhuwan Dhingra \qquad Partha Talukdar \\
  Google Research \\
  \texttt{\{jrcole,aditichaud,bdhingra,partha@google.com\}}}

\begin{document}
\maketitle
\begin{abstract}
Salient Span Masking (SSM) has shown itself to be an effective strategy to improve closed-book question answering performance. SSM extends general masked language model pretraining by creating additional unsupervised training sentences that mask a single entity or date span, thus oversampling factual information.
Despite the success of this paradigm, the span types and sampling strategies are relatively arbitrary and not widely studied for other tasks. 
Thus, we investigate SSM from the perspective of temporal tasks, where learning a good representation of various temporal expressions is important. To that end, we introduce Temporal Span Masking (TSM) intermediate training.
First, we find that SSM alone improves the  downstream performance on three temporal tasks by an avg. +5.8 points. 
Further, we are able to achieve additional improvements (avg. +0.29 points) by adding the TSM task. These comprise the new best reported results on the targeted tasks. 
Our analysis suggests that the effectiveness of SSM stems from the sentences chosen in the training data rather than the mask choice: sentences with entities frequently also contain temporal expressions. Nonetheless, the additional targeted spans of TSM can still improve performance, especially in a zero-shot context.
\end{abstract}

\begin{figure*}[ht!]
    \centering
    \includegraphics[width=\textwidth]{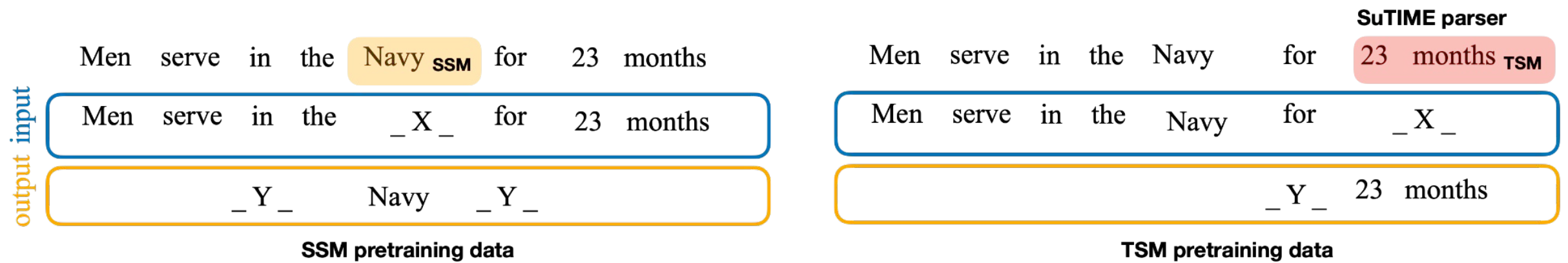}
    \caption{Overview of the TSM and SSM tasks: While SSM \cite{guu-etal-2020-realm} identifies named entities and dates as salient spans, for TSM we use \textsc{sutime} parser that captures other temporal expressions such as durations and intervals.
    \textsc{sutime} is first applied on raw sentences to identify any temporal expressions. Next, training data for TSM  is created from those sentences which have at least one temporal expression identified. In SSM, the masked spans comprise of named entity or date spans. Input to each task consists of the sentence with the selected span dropped out where the model is trained to predict the dropped tokens.
    }
    \label{fig:tsm_overview} 
\end{figure*}

\section{Introduction}
Salient Span Masking (SSM), first introduced by \citet{guu-etal-2020-realm} for retrieval-based language modeling, has shown performance gains for closed-book question answering (CBQA) \cite{roberts-etal-2020-much,ye2020studying}. 
SSM is a form of intermediate pretraining \cite{ye-etal-2021-influence}, where a pretrained model such as a BERT \cite{devlin-etal-2019-bert} or T5 \cite{raffel-etal-2020-exploring} is trained further before task-specific finetuning, generally on more specialized data that does not require expensive annotations.
Specifically, SSM uses the masked language modeling objective but only masks named entities and dates in sentences from English Wikipedia articles; these ``salient'' spans likely contain more facts, so the language model
must memorize more facts in order to do the task successfully \cite{petroni-etal-2019-language}.
The authors use a named entity recognition model to identify entity spans and a regular expression to identify date spans. 
While this works well for knowledge intensive downstream tasks,
such as entity-centric question answering, it remains unclear whether it is helpful for tasks
that are less aligned with the data, such as common sense  or temporal reasoning.
Moreover, is it possible to select spans that are more related to a downstream task in order to get further performance gains?

In this work, we investigate SSM for tasks that require understanding \emph{temporal} expressions.
While SSM does include dates, the tasks we investigate include other complex temporal expressions such as durations and intervals. 
To that end, we introduce Temporal Span Masking (TSM): an intermediate pretraining strategy for predicting spans that are likely temporal expressions
(\autoref{fig:tsm_overview}). 
Similar to SSM, TSM is automatically generated from English Wikipedia articles.
We compare models trained on TSM and SSM on three  temporal tasks, namely MC-TACO \cite{zhou-etal-2019-going}, TimeDIAL \cite{qin-etal-2021-timedial} and SituatedQA \cite{zhang-choi-2021-situatedqa}, and for one general-purpose question answering (QA) task of Natural Questions (NQ) \cite{nq}.
We summarize our contributions as follows:
\begin{itemize}
    \item We propose TSM Intermediate Training, which automatically selects temporal spans for masking.
    \item The new best reported results on the three temporal tasks: the best average performance is from a mixture of TSM and SSM. This mixture also does slightly better than SSM on Natural Questions. 
    \item Experiments investigating the role of different TSM and SSM span types, showing entity spans alone are helpful, which implies that difficult examples help improve representations of the unmasked spans as well.
\end{itemize}

\section{Methods}
Following \citet{roberts-etal-2020-much}, we utilize intermediate training to improve pretrained models' generalization to downstream tasks. 
All models are initialized from the encoder-decoder T5-1.1-XXL language model \citep{raffel-etal-2020-exploring}, which was shown to have the best closed-book QA performance in \citet{roberts-etal-2020-much}. 

\subsection{Background: Salient Span Masking}
Salient Span Masking (SSM) was first introduced by \citet{guu-etal-2020-realm} and is designed to specifically mask named entities and dates or \textit{salient spans}.
These salient spans are automatically identified from English Wikipedia using a Named Entity Recognition model to find entities as well as a regular expression to find dates. The authors mask one such span per sentence during training: the model must maximize the probability of the masked entity or date given the corrupted input sentence. \citet{guu-etal-2020-realm} designed the task to improve downstream performance on tasks requiring world knowledge in order to improve their retrieval-augmented model's ability to use retrieved texts. \citet{roberts-etal-2020-much} then

\subsection{Proposed: Temporal Span Masking}
\label{sec:tsm}
Inspired from the success of SSM, TSM is designed to address problems requiring temporal knowledge.
To create training examples for TSM, we automatically identify temporal expressions in a large corpus using \textsc{sutime} \cite{chang-manning-2012-sutime}, a rule-based temporal parser, that identifies temporal expressions from raw text. 
Given an input sentence, \textsc{sutime} is built to identify expressions of the following four types: \textbf{Time} which indicates a particular point in time such as \emph{next Monday}, \textbf{Duration} such as \emph{3 days}, \textbf{Set} which indicates periodic set of time that occur with some frequency such as \emph{every 4 years} and \textbf{Date} such as \emph{January 1}.

We run \textsc{sutime} on all of English Wikipedia\footnote{We use the 2020 snapshot of English Wikipedia (TFDS datadump \emph{wikipedia20201201en})}.
Specifically, we divide the articles into sentences, and apply \textsc{sutime}\footnote{\url{https://github.com/FraBle/python-sutime}} on each sentence.
For our TSM training data, we ensure that exactly one temporal span is masked per example.
So, if a sentence contains four temporal spans, we create four training examples with exactly one temporal span masked per example.
Details of the temporal distribution are in  \autoref{tab:tsmdata}. 
Each example is created by masking the tokens belonging to the temporal expression, as shown in  \autoref{fig:tsm_overview}, corrupting the input sentence by replacing the span with (\textunderscore X\textunderscore) and having the model predict the masked tokens.
The training objective is to maximize the probability of the target span given the corrupted input sentence, similar to T5's span corruption training objective \citep{raffel-etal-2020-exploring} and the SSM training objective \citep{roberts-etal-2020-much}.

\subsection{Model Variants}
All newly reported results are based on T5-1.1-XXL models. Proposed models are named for their intermediate training objective: TSM is trained solely on the masked temporal spans described above; SSM is trained solely on the training objective described in \citet{roberts-etal-2020-much}.\footnote{We note that the SSM-spans are derived from the 2018 snapshot of English Wikipedia (same as \citet{guu-etal-2020-realm}) while TSM-spans from the 2020 snapshot.}

We also investigate a version of the SSM pretraining data that only uses the \textit{named entity} spans identified by the NER model; in other words, all of the \textit{date} spans identified by the regular expression are removed, but the task is otherwise the same. We call this objective \textsc{Entities}. Finally, we compare against models that are trained proportional mixtures of both TSM+SSM and TSM+\textsc{Entities}. 

For baseline models, we use the same pretrained model (T5-1.1-XXL) with no intermediate training (T5), as well as a T5-1.1-XXL model which has been trained for an additional 100K steps on the prefix LM task (T5-LM; \citealt{lester-etal-2021-power}).

\subsection{Downstream Temporal Tasks}
We evaluate on three downstream temporal tasks for evaluation, finetuning a model on each task separately. 
Below, we briefly describe the tasks and datasets, with additional training details in \autoref{sec:training_appendix}.

\paragraph{MC-TACO} \citet{zhou-etal-2019-going} release a human-annotated dataset to measure temporal commonsense understanding.
It consists of 13k tuples of \emph{(sentence, question, candidate answer)} covering five types of common sense problems such as event frequency, event duration, event ordering, stationarity and event typical time.
Given a sentence context, a temporal question about that context, and a possible commonsense answer, the task is to determine whether the provided answer is reasonable for the given context.
For instance, for the event of \textit{taking a shower} with four possible answer choices \textit{five minutes, fifteen minutes, fifteen hours}, and \textit{fifteen years}, the first two are plausible and will have the \textit{yes} label while the latter two choices would be \textit{no}.
There is no training data released for this task, so we finetune the model on the provided validation set and evaluate on the test set.

\paragraph{TimeDIAL} \citet{qin-etal-2021-timedial} release a human-annotated multi-turn dialog dataset for measuring temporal commonsense understanding in a dialog setting. 
The dataset comprises of challenge test set with 1.1k dialog instances derived from the Daily Dialog dataset described in \citet{qin-etal-2021-timedial}.
TimeDIAL dialogs  mostly comprise of common sense instances where the answers generally consist of one temporal span. 
For instance, in the following dialog ``I'll just be a minute'., the span ``a minute'' may be masked out and the model is required to predict the masked span based on the dialog turns.
Given a dialog with a temporal expression masked out, the task is to correctly predict which two of the four provided answers are valid in the given context.
We report results without finetuning (styled TimeDIAL-0) as well as results from finetuning the model on the Daily Dialog dataset.
 
\paragraph{SituatedQA} \citet{zhang-choi-2021-situatedqa} release an open-domain QA dataset derived from existing question answering datasets with additional annotations that resolve temporal and geographic ambiguities.
Each example consists of a disambiguated question: for instance, ``Which COVID-19 vaccines have been authorized in the US [as of 2020]?'' or ``What was the first COVID-19 vaccine to be authorized [in the US]?''. 
For the purpose of this work, we focus on the temporal questions. These consist of 9K additional questions, with a training set of about 4.5K questions. 
We finetune on the training set and evaluate on the test set.
 
\subsection{Natural Questions}
While the focus of our method is improving temporal question answering performance, we also wanted to ensure that our method does not degrade performance on non-temporal question answering tasks. Thus, we also evaluate our model variants on Natural Questions \cite{nq}, using the ``open'' variant popularized by \citet{lee-etal-2019-latent}. These examples discard those questions without short answers or that require an evidence document to answer. These consist of about 87K questions for training and an additional 3.6k questions for validation, which we use for evaluation.
\begin{table*}[thb]
\begin{center}
\scalebox{0.9}{
\begin{tabular}{lcccccccccc}
 & \multicolumn{2}{c}{SituatedQA} & \multicolumn{2}{c}{MC-TACO} & \multicolumn{2}{c}{TimeDIAL} & \multicolumn{2}{c}{TimeDIAL-0} & &  \\
\cmidrule(lr){2-3} \cmidrule(lr){4-5} \cmidrule(lr){6-7} \cmidrule{8-9}
Model & F1 & EM & F1 & EM & 1-Best & 2-Best & 1-Best & 2-Best & Overall & Overall-0  \\
\midrule
\textsc{Best Reported} & -- & 18.53 & 82.92 & 63.81 & -- &  76.10 & -- & 50.60 & 52.74 & 44.31 \\
\midrule
\textsc{T5} & 25.75 & 19.78 & 84.00 & 64.56 & 99.91 & \textbf{84.50} & 90.85 & 37.59 & 56.28 & 40.64 \\
\textsc{T5-LM} & 25.38 & 19.63 & 81.99 & 59.83 & 99.91 & 80.60 & 86.87 & 32.16 & 53.35 & 37.21  \\
\textsc{SSM} & 29.92 & 23.12 & 85.88 & 68.39 & 99.73 & 84.06 & 96.74 & 67.21 & 58.52 & 52.91 \\
\midrule
\textsc{Entities} & 29.42 & 22.82 & 85.47 & 66.59 & 99.91 & 83.06 & 97.64 & 67.93 & 57.49 & 52.45 \\
\textsc{TSM} & 27.42 & 21.18 & 84.89 & 65.92 & 99.91 & 83.88 & \textbf{99.82} & \textbf{77.54} & 56.99 & 54.88  \\
\textsc{TSM+SSM} & 29.33 & 22.76 & \textbf{86.20} & 67.64 & \textbf{100.0} & 83.78 & 98.19 & 73.10 & 58.03 & 54.5  \\
\textsc{Entities+TSM} & \textbf{30.78} & \textbf{24.60} & 85.32 & \textbf{68.47} & 99.91 & 84.24 & 98.91 & 76.09 & \textbf{59.09} & \textbf{56.39} \\
\bottomrule
\end{tabular}
}
\caption{Aggregate metrics across the three datasets. \textit{Overall} performance is the simple arithmetic average of the harder metric for each approach (EM, EM, 2B); \textit{Overall-0} uses TimeDIAL-0 instead of TimeDIAL. The second section contains our runs of earlier models; the Best Reported uses best known published numbers. The third section represents our models. Note that all models (in the second and third sections) are based on T5-1.1-XXL models. Best Reported results are ALBERT \citep{lan2019albert} from \citet{abramson2022application} for TimeDIAL, BART results from \citet{zhang-choi-2021-situatedqa}, and DeBERTa \citep{he-etal-2020-deberta} results from the leaderboard for MC-TACO. Note that F1 for MC-TACO is based on the precision/recall over answers and EM is based on labeling every answer for a question correctly, while the F1 for SituatedQA is based on the token-level F1 of the answer span.}
\label{tab:overall}

\end{center}
\end{table*}
\section{Results and Discussion}
Our main results can be found in \autoref{tab:overall}. Results including Natural Questions can be found in \autoref{tab:natural_questions}. Note that the Natural Questions results have minor variations from published numbers; we ran these baselines ourselves, and it is possible the training setup differed slightly.

\begin{table}[ptbh]
\begin{center}
\scalebox{0.8}{
\begin{tabular}{lcccc}
\toprule
Model & F1 & EM & Overall & Overall-0 \\
\midrule
\textsc{SSM} & 41.57 & 34.6 & 52.54 & 48.33 \\
\textsc{T5} & 39.35 & 32.38 & 50.31 & 38.58  \\
\textsc{T5-LM} & 37.16 & 31.14 & 47.80 & 35.69 \\
\midrule
\textsc{Entities} & 41.21 & 34.52 & 51.75 & 47.97 \\
\textsc{TSM} & 39.24 & 32.69 & 50.92 & 49.33 \\
\textsc{TSM+SSM} & 41.80 & 35.10 & 52.3 & 49.65  \\
\textsc{Entities+TSM} & \textbf{41.89} & \textbf{35.18} & \textbf{53.11} & \textbf{51.09} \\
\bottomrule
\end{tabular}
}
\caption{Results on Natural Questions -- \emph{Overall} and \emph{Overall-0} results include the same metrics from \autoref{tab:overall} with Natural Questions (EM) included. The first section represents our baselines. Note that all models are based on T5-1.1-XXL models.
}
\label{tab:natural_questions}

\end{center}
\end{table}

\paragraph{T5 and T5-LM} The T5 model sets a relatively high baseline compared to previously reported models. The T5-LM model's extra non-domain-specific pretraining does not help on any task, suggesting extra training steps does not in of itself cause improvements on these tasks.

\paragraph{Entities} The \textsc{Entities} model, which is trained on only non-temporal entity spans, performs better overall than the TSM task. It only does worse on the TimeDIAL dataset, which is almost entirely focused on conversational, non-knowledge based contexts. It still does substantially better than the base T5 model when no finetuning data is available. This high performance is possibly due to the prevalence of temporal spans in the SSM training data. Running \textsc{sutime} on the \textsc{Entities} data reveals that 45\% of its training examples contain at least one date, duration, set, or time.

This suggests that sentences with named entities in general already carry temporal-salient information useful for downstream temporal tasks. See \autoref{app:discussion_appendix} for a full breakdown of the co-occurrences.

\paragraph{SSM} The SSM model is the second best overall. It benefits from both its own date spans as well as the frequent presence of temporal spans in the entities data, suggesting difficult example sentences are more important than the type of masked span. It does worse on TimeDIAL-0, however, where the task is to score the best temporal span.

\paragraph{TSM} The TSM model improves upon the baseline T5 model but is worse overall than the SSM model. However, it is the best on TimeDIAL-0. This is likely because the DailyDialog training dataset is relatively large, which may overcome the need for intermediate pretraining altogether. Note that TSM achieves a mild performance improvement over the baseline T5 model on Natural Questions, but is notably worse than the other intermediate training methods. 

\paragraph{TSM+SSM} The TSM+SSM model improves over TSM but is worse than SSM outside of TimeDIAL-0. One possible reason for the regression is that TSM and SSM have overlapping Date span examples, which may make the intermediate task easier and thus less useful. However, it is slightly better than SSM on Natural Questions.

\paragraph{Entities+TSM} The \textsc{Entities+TSM} model performs the best overall: with and without the extra training data for TimeDIAL. It has the benefit of TSM spans without containing overlapping spans or losing the world knowledge from entity spans. It also performs slightly better than SSM on Natural Questions.

\begin{table}[t!]
\begin{center}
\begin{tabular}{lccccc}
\toprule
Model & \textsc{T5} & \textsc{SSM} & \textsc{E+TSM} \\  
\midrule
Duration & 88.64 & 88.28 & \textbf{89.09} \\
Set & 86.39 & \textbf{88.22} & 87.79 \\
Time & 87.74 & \textbf{88.70} & 87.70 \\
Date & 65.75 & 66.63 & \textbf{66.87} \\
Entities & 42.17 & 45.76 &\textbf{46.54} \\
\bottomrule
\end{tabular}
\caption{Aggregated performance across types for baselines and the best overall model (\textsc{Entities}+TSM). For TimeDIAL, each answer span is labeled by \textsc{sutime}. Duration includes MC-TACO's ``Event Duration''; Set includes MC-TACO's ``Frequency'', Time and Date both include MC-TACO's ``Typical Time''; Date also includes SituatedQA; Entities include all of MC-TACO and SituatedQA. Note that the majority of the gains from SSM/TSM seem to be from Entities and Dates. See \autoref{app:discussion_appendix} for rationale behind these choices.
}
\label{tab:breakdown}

\end{center}
\end{table}

\paragraph{By Type} We analyze model performance by temporal type in \autoref{tab:breakdown}. The main improvement of both \textsc{SSM} and \textsc{Entities+TSM} is in entity and date tasks. Surprisingly, TSM shows a regression on time tasks, and only gets a slight improvement on duration tasks. One possible hypothesis for this is that temporal expressions may be more informative when co-occurring with an entity. Note that these numbers are based on the trained versions of each dataset, excluding Natural Questions. Note that SituatedQA contains further breakdowns based on the scope of the date, but this does not map well to the other datasets.

\section{Related Work}
\paragraph{Span Masking and Intermediate Training}
Salient Span Masking \citep{guu-etal-2020-realm} came out of a series of efforts like SpanBERT \citep{joshi-etal-2020-spanbert} to select more difficult examples to improve models memorization of the text.

Most similar to us, \citet{ye-etal-2021-influence} explore a similar paradigm of choosing better spans for a downstream task (e.g., entity linking or relation extraction) where they experiment with both a heuristic masking policy similar to SSM and also a learned masking policy.
They similarly find that masking spans that resemble  downstream tasks improve performance, however, they also note that learned masking policies suffer from overfitting.
\citet{yang-etal-2020-improving} and \citet{zhou-etal-2020-temporal} explore intermediate training by designing heuristics to identify sentences containing temporal expressions and then adding additional tasks and losses, rather than using span masking. TSM differs in more closely resembling the pretraining task.

\citet{levine2021pmimasking} use pointwise mutual information to jointly mask highly correlated spans to avoid the model relying on local signals but rather learning from the broad context.
They find this leads to faster and better pretraining.
In the future, it might be interesting to see how PMI-spans can combine with knowledge-oriented span techniques such as SSM, TSM, and whether they can help in the intermediate training paradigm.

\paragraph{Temporal Understanding}
There has been a surge of interest in probing models' temporal awareness. While we evaluate on a three tasks, it is far from an exhaustive evaluation and we leave further evaluations of our method to future work.

Recently,
\citet{thukral-etal-2021-probing} and \citet{vashishtha-etal-2020-temporal} construct NLI datasets to test whether pretrained models understand certain types of common sense temporal expressions, such as containment. 
To probe common sense, we use TimeDIAL \citep{qin-etal-2021-timedial} for its naturalistic dialogues as well as MC-TACO \citep{zhou-etal-2020-temporal}, which uses a diverse set of situations and temporal expressions. 

For factual questions, 
open-response temporal questions are closely aligned with our work (e.g., TimeQA; \citealt{chen2021dataset}; TempLAMA; \citealt{dhingra2022time}). 
All of TempLAMA, TimeQA, and SituatedQA \citep{zhang-choi-2021-situatedqa} rely primarily on the year as the main temporal expression being tested, where facts are scoped to the provided years. To probe temporally scoped facts, we use SituatedQA for its more naturalistic questions.

\section{Conclusion}
In this work, we investigate SSM as it relates to temporal tasks that require understanding both commonsense and world knowledge questions and propose a new intermediate training method which selects spans generated by a temporal parser. 
These intermediate training strategies result in the best overall reported results on the selected downstream tasks. 
However, we find that even the entity spans from SSM are helpful for temporal tasks, likely because entity-containing examples also contain informative temporal knowledge.
\section*{Limitations}
This analysis investigates only the encoder-decoder model architecture: in particular, encoder-only models such as BERT \citep{devlin-etal-2019-bert} and decoder-only models such as GPT-2 \citep{radford2019language} are excluded. Further, large language models, such as PaLM \citep{chowdhery2022palm} or GPT-3 \citep{brown2020language} are also not investigated. See \autoref{app:discussion_appendix} for further discussion.

\section*{Acknowledgements}
We would like to thank Kelvin Guu and Srini Narayanan, as well as our anonymous reviewers, for their helpful feedback on a previous version of this manuscript.

\bibliography{anthology,custom}
\bibliographystyle{acl_natbib}

\clearpage

\appendix

\section{Training Details}
\label{sec:training_appendix}
All models are initialized from the public T5-1.1-XXL checkpoints.\footnote{https://github.com/google-research/text-to-text-transfer-transformer}

\subsection{Intermediate Training}
\label{sec:intermediate}
We use 256 Cloud TPU v3 cores for the intermediate training procedure using a batch size of 2048 and the fixed default learning rate of 0.001. Training generally proceeds for one epoch, which is between 100-150K steps depending on the precise task, though we used early stopping for the TSM and TSM+SSM models based on MC-TACO performance, as they seem to overfit.

\subsection{Finetuning}
\label{sec:finetuning}
We use 64 Cloud TPU v3 cores for finetuning and inference on all tasks. For MC-TACO, Natural Questions, and SituatedQA, we use the same fixed learning rate of 0.001 and train for 10K steps with batch size 128. For TimeDIAL, we attempt to follow their training setup more closely, and use a lower learning rate (0.0001) and train for up to 100K steps, still with batch size 128 and use early stopping on the validation set to inform when to stop. For most of the models that had intermediate training, the early stopping point was for 10K steps. However, for the basic T5 model, it was after 20K steps (improving from ($82.97 \rightarrow 84.51$)), implying that it can overcome its lack of intermediate training with additional finetuning data. Note that in the zero-shot variant, no finetuning is done.

\section{Further Discussion}
\label{app:discussion_appendix}
\subsection{Other Experiments}
We previously experimented with the T5~Large and T5-XL models, as well the as 1.0 versions of the T5 models that were first described in \citet{raffel-etal-2020-exploring}. In general, larger models and the 1.1 versions worked better. While we refrain from reporting results due to inconsistent setups, in general the smaller models were notably worse, such that distinguishing between two similar setups (such as TSM and SSM) was difficult on many tasks. While we know of no work testing salient span masking on extremely large models, it is possible it would actually show a larger impact, based on this trend. While left-to-right decoding serves as an awkward fit for the paradigm, if our hypothesis on the reason why SSM works is correct, then it should not prove to be a substantial hurdle. See also below for more discussion on said hypothesis.

\begin{table}[t!]
\begin{center}
\begin{tabular}{lc}
\toprule
Temporal Type & Number of sentences\\  
\midrule
Date & 56,520,912 \\
Duration & 8,182,819 \\
Set & 1,797,929 \\
Time & 2,281,198 \\
\bottomrule
\end{tabular}
\caption{TSM data statistics: The above table describes the distribution of temporal spans in the English Wikipedia data, which comprises of 121M sentences. 
}
\label{tab:tsmdata}

\end{center}
\end{table}
\begin{table}[t!]
\begin{center}
\begin{tabular}{lc}
\toprule
Span Type & Number of sentences\\  
\midrule
Entity & 78,139,341 \\
Date & 32,023,769 \\
\bottomrule
\end{tabular}
\caption{SSM data statistics: The above table describes the distribution of salient spans in the Wikipedia data as processed by \cite{guu-etal-2020-realm}, which comprises of 82M sentences. Each row denotes the number of sentences that contain at least one of the respective span.
}
\label{tab:ssmdata}

\end{center}
\end{table}
\begin{table}[t!]
\begin{center}
\begin{tabular}{lcc}
\toprule
TSM Span Type & \multicolumn{2}{c}{SSM Span Type}  \\  
& Named Entity & Date \\
\midrule
Date & 29,771,242 & - \\
Duration & 5,159,229 & 3,144,592\\
Set & 1,226,333 & 844,222\\
Time & 915,084 & 411,436\\
\bottomrule
\end{tabular}
\caption{We apply \textsc{SUTIME} on the SSM training data \cite{guu-etal-2020-realm} to investigate how many sentences contain temporal information. Each column denotes the number of sentences that contain the SSM identified span (e.g. \textit{named entity} or \textit{date})  and each row denotes the number of those sentences in which \textsc{SUTIME} identified the corresponding temporal span.
Number of sentences with at least one \textit{named entity}: 78,139,341 \\
Number of sentences with at least one \textit{date}: 32,023,769 (Table \ref{tab:ssmdata})
}
\label{tab:ssmdatatsmsutime}
\end{center}
\end{table}

\subsection{Span Distribution vs. Text Distribution}
Our hypothesis for SSM's effectiveness is due to it oversampling difficult sentences. This is based on the performance gain for the \textsc{Entities} intermediate training as well as the number of temporal spans that occur in the SSM training data.
\autoref{tab:ssmdatatsmsutime} shows the results of \textsc{SUTIME} parser on the SSM training data, and as we can see, significant portion of the SSM data (45\%) has temporal spans.
 \autoref{tab:ssmdatatsmsutime} shows the breakdown of different temporal spans for each SSM salient span type.
 We leave an exact test of this for future work, but if this is true, then we might expect left-to-right decoding models to also benefit from the sampling procedure of SSM, even though they do not use a masked language modeling paradigm for training.

\subsection{Span Types in Table \ref{tab:tsmdata}}\label{sec:table2}
Mapping MC-TACO's span types is somewhat helpful to see the performance breakdown. Note that these are now based on individual answers, while MC-TACO's strict match metric is based on correctly labeling all answers for a given question.

\paragraph{Entities} While MC-TACO is a common sense dataset, it frequently relies on reasoning about relatively complicated phenomena. While it is common sense to know that a dynasty does not rule in China for only a few minutes, it is still required to know more about China and dynasties to answer the question correctly. TimeDIAL on the other hand is normally ordinary conversations that are not very entity-centric. SituatedQA is derived from Natural Questions, which is an information seeking dataset that frequently features entities.

\paragraph{Duration} MC-TACO's event duration maps well to the Duration type in \textsc{sutime}. While there may be some SituatedQA examples that include durations, we do not filter for them.

\paragraph{Set} MC-TACO's Frequency type asks question of the "How often" nature while sets frequently have answer types of that nature e.g., "every third sunday", but this is not a perfect mapping.

\paragraph{Date} MC-TACO's typical time sometimes includes dates, but it is less likely to be a specific date and more likely to be a generic date like Sunday, rather than a specific knowledge-based date. SituatedQA questions always include dates that decontextualize Natural Questions. 

\paragraph{Time} MC-TACO's typical time sometimes corresponds with times as well, but they are again less likely to be specific. Unfortunately, Date and Time are not separated in MC-TACO. 

\paragraph{Other MC-TACO Types} Note that we did not include the ``Stationarity'' or the ``Event Ordering'' MC-TACO types in the breakdown, as they do not correspond well to any \textsc{sutime} type.

\end{document}